\documentclass{svproc}
\usepackage{url}
\usepackage{graphicx}
\usepackage{booktabs}
\usepackage{amsmath}
\usepackage{bm} 

\begin{document}
\mainmatter              

\title{Enhancing Knowledge Graph Construction: Evaluating with Emphasis on Hallucination, Omission, and Graph Similarity Metrics}

%
\titlerunning{Enhancing Knowledge Graph Construction Evaluation}  
%
\author{Hussam Ghanem\inst{1} \and Christophe Cruz\inst{1}}
%
%
%
\institute{ICB, UMR 6306, CNRS, Université de Bourgogne, 21000 Dijon, France\\
\texttt{https://icb.u-bourgogne.fr/}
}
\maketitle              

\begin{abstract}
Recent advancements in large language models have demonstrated significant potential in the automated construction of knowledge graphs from unstructured text. This paper builds upon our previous work \cite{nous}, which evaluated various models using metrics like precision, recall, F1 score, triple matching, and graph matching, and introduces a refined approach to address the critical issues of hallucination and omission. We propose an enhanced evaluation framework incorporating BERTScore for graph similarity, setting a practical threshold of 95\% for graph matching. Our experiments focus on the Mistral model, comparing its original and fine-tuned versions in zero-shot and few-shot settings. We further extend our experiments using examples from the KELM-sub training dataset, illustrating that the fine-tuned model significantly improves knowledge graph construction accuracy while reducing the exact hallucination and omission. However, our findings also reveal that the fine-tuned models perform worse in generalization tasks on the KELM-sub dataset. This study underscores the importance of comprehensive evaluation metrics in advancing the state-of-the-art in knowledge graph construction from textual data.
\end{abstract}
\section{Introduction}
Knowledge Graphs (KGs) play a crucial role in organizing complex information across diverse domains, such as question answering, recommendations, semantic search, etc.  However, the ongoing challenge persists in constructing them, particularly as the primary sources of knowledge are embedded in unstructured textual data such as press articles, emails, and scientific journals. This challenge can be addressed by adopting an information extraction approach, sometimes implemented as a pipeline. It involves taking textual inputs, processing them using Natural Language Processing (NLP) techniques, and leveraging the acquired knowledge to construct or enhance the KG.

In-context learning, as discussed by \cite{min2022rethinking}, coupled with prompt design, involves telling a model to execute a new task by presenting it with only a few demonstrations of input-output pairs during inference. Instruction fine-tuning methods, exemplified by InstructGPT \cite{ouyang2022training} and Reinforcement Learning from Human Feedback (RLHF) \cite{stiennon2020learning}, markedly enhance the model's ability to comprehend and follow a diverse range of written instructions. Numerous large language models (LLMs) have been introduced in the last year, as highlighted by \cite{mihindukulasooriya2023text2kgbench}, particularly within the ChatGPT \cite{openai2023gpt} like models, which includes GPT-3 \cite{brown2020language}, LLaMA \cite{touvron2023llama}, Mistral \cite{jiang2023mistral}, and Starling \cite{zhu2023starling}. These models can be readily repurposed for KG construction from text by employing a prompt design that incorporates instructions and contextual information.

The task of converting textual information into structured KGs has gained significant traction with the advent of LLMs. These models offer unprecedented capabilities in understanding and generating human-like text, making them invaluable for a variety of NLP applications. Our previous work \cite{nous} explored different approaches to the Text-to-Knowledge Graph (T2KG) construction task, including Zero-Shot Prompting (ZSP) \cite{carta2023iterative}, Few-Shot Prompting (FSP) \cite{han2023pive}, and Fine-Tuning (FT) \cite{ershov2023case} of LLMs, employing models such as Llama2 \cite{touvron2023llama}, Mistral \cite{jiang2023mistral}, and Starling \cite{zhu2023starling}. In this work, we will include a little state of the art on contributions that use these three approaches (Section \ref{sec:2}).

While traditional metrics like precision, recall, F1 score, triple matching, and graph matching provide a baseline for evaluating these models, they often overlook critical qualitative aspects of the generated graphs, such as hallucinations (incorrect or spurious triples) and omissions (missing relevant triples). Addressing these gaps, our current study introduces a refined evaluation framework that incorporates refined hallucination and omission metrics, and also incorporates BERTScore to measure the similarity between generated and ground truth graphs, setting an 95\% similarity threshold for graph matching. This nuanced approach aims to provide a more comprehensive assessment of the models' performance in generating accurate and complete knowledge graphs.

In this paper, we specifically focus on comparing the original Mistral model and our finetuned Mistral (from our previous work) under zero-shot and few-shot settings. Additionally, we extend our experiments to include the KELM-sub dataset, utilizing few-shot examples to demonstrate that fine-tuning on a specific domain (WebNLG) significantly enhances performance when applied to related but distinct datasets with just few examples.

The present study is organized as follows, Section \ref{sec:2} presents a comprehensive overview of the current state-of-the-art approaches for Text to KG (T2KG) Construction and its evaluation metrics. In the Section \ref{sec:3}, we present the general architecture of our proposed implementation (method), with datasets, metrics, and experiments. Section \ref{sec:4} then encapsulates the findings and discussions, presenting the culmination of results. Finally, Section \ref{sec:5} critically examines the strengths and limitations of these techniques.

\section{Background}
\label{sec:2}

The current state of research on knowledge graph construction using LLMs is discussed. Three main approaches are identified: Zero-Shot, Few-Shot, and Fine-Tuning. Each approach has its own challenges, such as maintaining accuracy without specific training data or ensuring the robustness of models in diverse real-world scenarios. Evaluation metrics used to assess the quality of constructed KGs are also discussed, including semantic consistency and linguistic coherence. This section highlight methods and metrics to construct KGs and evaluate the result.

\subsection{Zero Shot}
\label{sec:2.1.} 
Zero Shot methods enable KG construction without task-specific training data, leveraging the inherent capabilities of LLMs. \cite{carta2023iterative} introduce an innovative approach using LLMs for knowledge graph construction, employing iterative zero-shot prompting for scalable and flexible KG construction. \cite{zhu2023llms} evaluate the performance of LLMs, specifically GPT-4 and ChatGPT, in KG construction and reasoning tasks, introducing the Virtual Knowledge Extraction task and the VINE dataset, but they do not take into account open sourced LLMs as LLaMA \cite{touvron2023llama}. \cite{bi2024codekgc} address the limitations of existing generative knowledge graph construction methods by leveraging large generative language models trained on structured data. The most of these approaches having the same limitation, which is the use of closed and huge LLMs as ChatGPT or GPT4 for this task. Challenges in this area include maintaining accuracy without specific training data and addressing nuanced relationships between entities in untrained domains.

\subsection{Few Shot}
\label{sec:2.2.}
Few Shot methods focus on constructing KGs with limited training examples, aiming to achieve accurate knowledge representation with minimal data. \cite{han2023pive} introduce PiVe, a framework enhancing the graph-based generative capabilities of LLMs, and the authors create a verifier which is responsable to verifie the results of LLMs with multi-iteration type. \cite{chen2023knowledge} investigate LLMs' application in relation labeling for e-commerce Knowledge Graphs (KGs). As ZSP approaches, FSP approaches use closed and huge LLMs as ChatGPT or GPT4 \cite{openai2023gpt} for this task. Challenges in this area include achieving high accuracy with minimal training data and ensuring the robustness of models in diverse real-world scenarios.

\subsection{Fine-Tuning}
\label{sec:2.3.} 
Fine-Tuning methods involve adapting pre-trained language models to specific knowledge domains, enhancing their capabilities for constructing KGs tailored to particular contexts. \cite{ershov2023case} present a case study automating KG construction for compliance using BERT-based models. This study emphasizes the importance of machine learning models in interpreting rules for compliance automation. \cite{yang2023chatgpt} propose Knowledge Graph-Enhanced Large Language Models (KGLLMs), enhancing LLMs with KGs for improved factual reasoning capabilities. These approaches that applied FT, they do not use new generations of LLMs, specially, decoder only LLMs as Llama, and Mistral. Challenges in this domain include ensuring the scalability, interpretability, and robustness of fine-tuned models across diverse knowledge domains.

\subsection{Evaluation metrics}
\label{sec:2.4.}
As we employ LLMs to construct KGs, and given that LLMs function as Natural Language Generation (NLG) models, it becomes imperative to discuss NLG criteria. In NLG, two criteria \cite{ferreira2019neural} are used to assess the quality of the produced answers (triples in our context).

The first criterion is semantic consistency or Semantic Fidelity, which includes:
\begin{itemize}
\item \textbf{Hallucination}: Presence of information (facts) in the generated text that is absent in the input data.
\item \textbf{Omission}: Omission of information present in the input data from the generated text.
\item \textbf{Redundancy}: Repetition of information in the generated text (not considered in our evaluation).
\item \textbf{Accuracy}: Exact match between the input and generated text without modification.
\item \textbf{Ordering}: Sequence of information in the generated text differing from the input data (not considered in our evaluation).
\end{itemize}

The second criterion is linguistic coherence or Output Fluency, which evaluates the fluidity and linguistic correctness of the generated text. This criterion is not considered in our evaluation.

In their experiments, \cite{mihindukulasooriya2023text2kgbench} calculated three hallucination metrics - subject hallucination, relation hallucination, and object hallucination - using preprocessing steps like stemming. They used the ground truth ontology and test sentence to determine if an entity or relation is present, considering any disparity between them as hallucination.

The authors of \cite{han2023pive} evaluated their experiments using several metrics, including Triple Match F1 (T-F1), Graph Match F1 (G-F1), G-BERTScore (G-BS) from \cite{saha2021explagraphs}, and Graph Edit Distance (GED) from \cite{abu2015exact}. The GED metric measures the distance between the predicted and ground-truth graphs by calculating the number of edit operations needed to transform one into the other. To adhere to the semantic consistency criterion, we use the terms "omission" and "hallucination" instead of "addition" and "deletion," respectively.

\section{Propositions}
\label{sec:3}

\begin{figure}[ht]
\centering
  \includegraphics[width=0.99\textwidth]{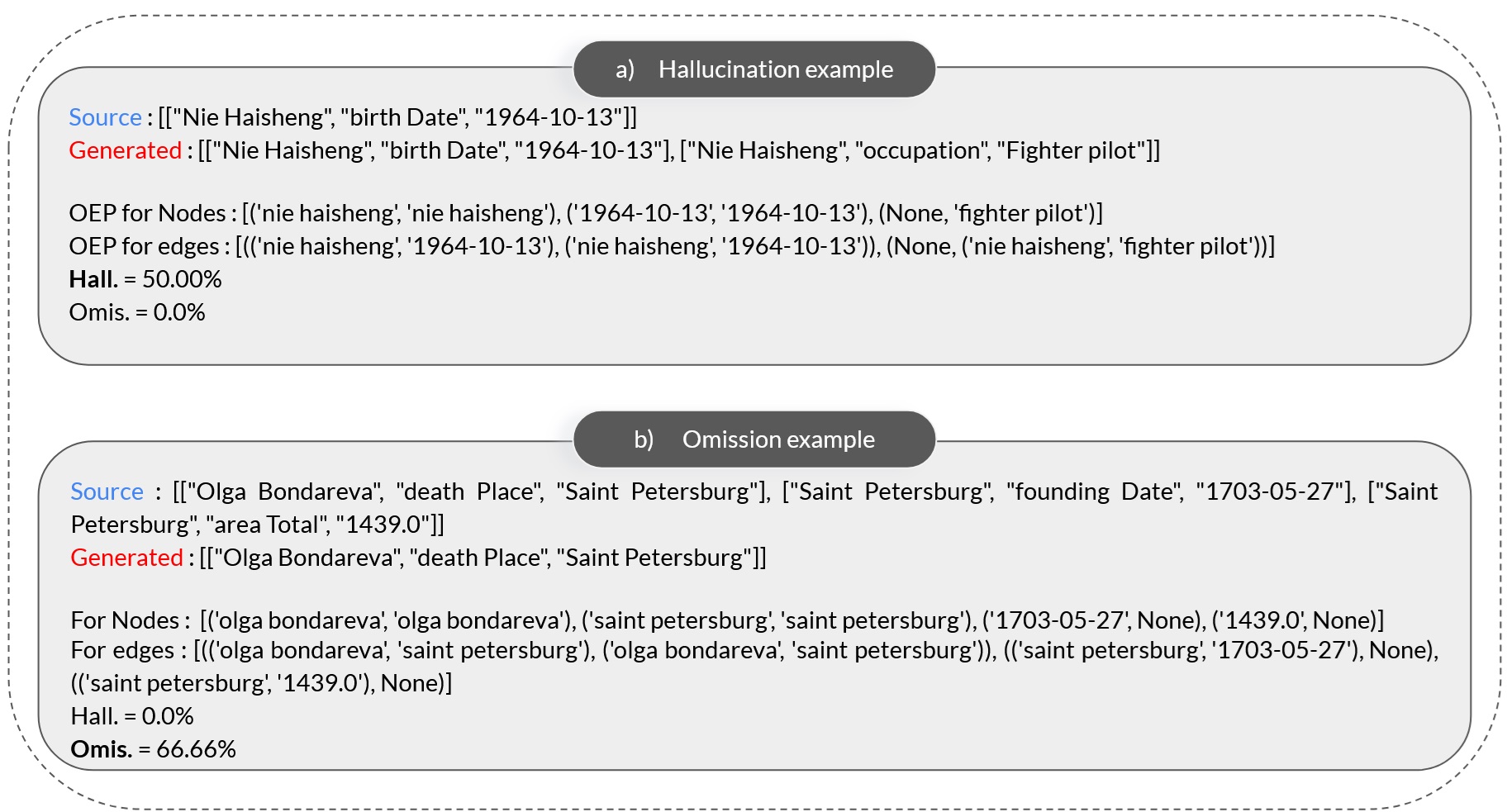}
\caption{Results examples}
\label{fig:3}       
\end{figure}

This section outlines our approach to evaluate the quality of generated KGs using metrics like T-F1, G-F1, G-BS, and GED. We also discuss the use of Optimal Edit Paths (OEP) to determine the precise number of operations needed to transform the predicted graph into an identical representation of the ground-truth graph. This method helps in calculating omissions and hallucinations in the generated graphs. Unlike our previous work where we marked a single hallucination or omission per generated graph, we now calculate the exact number of hallucinations and omissions for each generated graph (Fig. \ref{fig:3}. Previously, we used examples from the WebNLG+2020 dataset \cite{gardent2017webnlg} for testing with FSP techniques and trained LLMs using the FT technique. In this work, we take the best fine-tuned model (Mistral) from our previous work and apply zero/few-shot learning, comparing it with the original Mistral. Examples for few-shot learning are taken from WebNLG+2020 and the KELM-sub training dataset, and inference is applied on both datasets. We then compare these results with our previous work where models were applied on WebNLG+2020 and KELM-sub using examples from the WebNLG+2020 training dataset.

\begin{figure}[ht]
\centering
  \includegraphics[width=\linewidth]{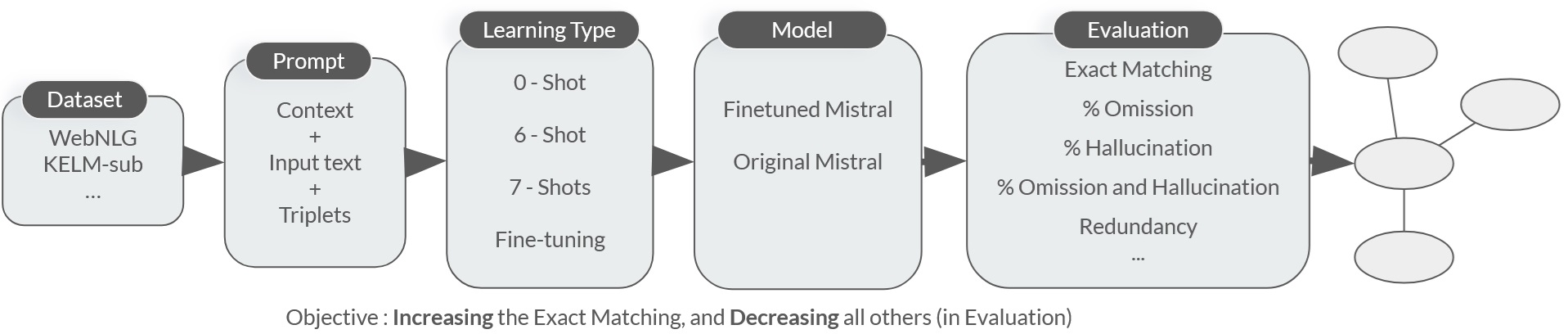}
\caption{Overall experimentation's process}
\label{fig:1}       
\end{figure}

\subsection{Overall experimentation's process}
\label{sec:3.1.}
In our previous work, we leveraged the WebNLG+2020 and KELM-sub datasets, specifically the version curated by \cite{han2023pive}. Their preparation of graphs in lists of triples proves beneficial for evaluation purposes. We utilize these lists and employ NetworkX \cite{hagberg2008exploring} to transform them back into graphs, facilitating evaluations on the resultant graphs. This step is instrumental in performing ZSP, FSP, and FT LLMs on these datasets. In this work, we will use examples from the training dataset of KELM-sub to do few-shot learning on the original and the finetuned (from our previous work) Mistral model.

Fig. \ref{fig:1} illustrates the different stages of our experimentation process, including data preparation, model selection, training, validation, and evaluation. The process begins with data preparation, where the WEBNLG dataset is preprocessed and split into training, validation, and test sets. Next, the learning type is selected, and different models are trained using the training set. The trained models are then evaluated on the validation set to evaluate their performance. Finally, the best-performing model is selected and validated on the test set to estimate its generalization ability.

\subsection{Prompting learning}
\label{sec:3.2.}

\begin{figure}[ht]
  \centering
  \includegraphics[width=\linewidth]{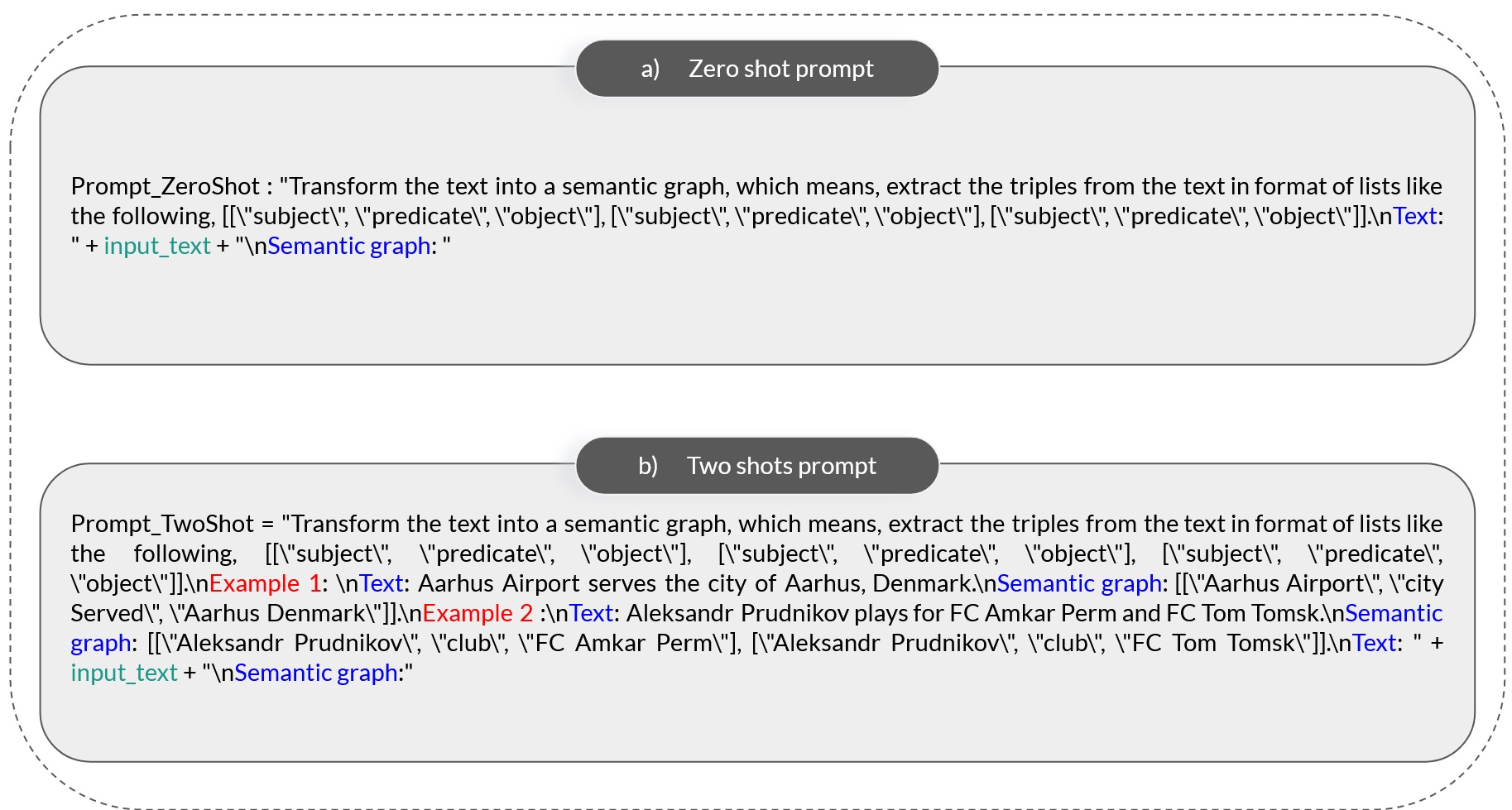}
  \caption{Prompting examples}
\label{fig:2}       
\end{figure}

In this phase, we use ZSP and FSP techniques on LLMs to evaluate their proficiency in extracting triples for KG construction. We merge examples from the KELM-sub test dataset with our adapted prompt, strategically modified for contextual guidance without a support ontology description, as demonstrated by \cite{mihindukulasooriya2023text2kgbench}. The prompts for ZSP and FSP are shown in Fig. \ref{fig:2}(a) and Fig. \ref{fig:2}(b).

For ZSP, we started with the method from \cite{han2023pive}, using the directive "Transform the text into a semantic graph" and enhanced it with additional sentences for our LLMs (Fig. \ref{fig:2}(a)). For FSP, we used 6-shot learning, corresponding to the maximum KG size in KELM-sub, feeding the prompt with six examples of varying sizes (Fig. \ref{fig:2}(b)).

\subsection{Postprocessing}
\label{sec:3.4.}

To evaluate the generated KGs against ground-truth KGs, we clean the LLM outputs by transforming generated graphs into organized lists of triples and transferring them to textual documents. This rule-based processing removes corrupted text outside the lists of triples, optimizing our evaluation process for metrics like G-F1, GED, and OEP (Section \ref{sec:3.6.}).

In our previous work, instructing LLMs to produce lists of triples sometimes resulted in unstructured text, which we addressed by substituting the generated text with an empty list of triples ('[["","",""]]'). This approach, however, underestimated hallucinations. In the current work, as illustrated in Fig. \ref{fig:3}, we calculate the exact hallucination and omission for each generated graph through qualitative evaluation of two randomly generated graphs.

\subsection{Experiment's evaluation}
\label{sec:3.6.}

To evaluate the generated graphs against ground-truth graphs, we use metrics such as T-F1, G-F1, G-BS \cite{saha2021explagraphs}, and GED \cite{abu2015exact} as in \cite{han2023pive}. We also use Optimal Edit Paths (OEP) to calculate omissions and hallucinations in the generated graphs.

Our evaluation follows \cite{han2023pive}'s methodology, especially in computing GED and G-F1, and involves constructing directed graphs from lists of triples using NetworkX \cite{hagberg2008exploring}. Unlike \cite{mihindukulasooriya2023text2kgbench}, we do not use the ground truth test sentence of an ontology. Instead, we assess omissions and hallucinations using OEP, which provides the precise path of the edit, allowing exact quantification of these errors.

For example, Fig. \ref{fig:3} shows 2 omissions ('b)') and 1 hallucination 'a)' in using one of two paths "OEP for nodes" or OEP for edges". Previously, we incremented the global hallucination metric for all graphs if $\geq$ 1 hallucinations or omissions were found. In the current work, we use OEP to detect the exact percentage of hallucination or omission in a generated graph, experimenting on 2 random examples from the WebNLG+2020 test dataset (Fig. \ref{fig:3}).

Different from our previous work, our experiments are evaluated using examples from the KELM-sub test dataset (Table \ref{tab:tab_res_kelm} and Table \ref{tab:tab_res}). Our primary goal is to improve G-F1, T-F1, G-BS  and GM-GBS metrics, while reducing GED, hallucination, and omission.

\subsection{Mathematical representation of the used metrics}
\label{sec:3.7.}

This study refines the metrics used for evaluating hallucinations and omissions in generated graphs and introduces a new metric, Graph Matching using Graph BERTScore (GM-GBS). In our previous work, we detailed the mathematical representation of all metrics used.

The G-BS metric evaluates graph matching by treating edges as sentences and using BERTScore to measure alignment between predicted and ground-truth edges. The F1 score for G-BS is calculated as follows:
\begin{align*}
R_{\text{BERT}} &= \frac{1}{|x|} \sum_{x_i \in x} \max_{\hat{x}_j \in \hat{x}} x_i^T \hat{x}_j, \\
P_{\text{BERT}} &= \frac{1}{|\hat{x}|} \sum_{\hat{x}_j \in \hat{x}} \max_{x_i \in x} x_i^T \hat{x}_j, \\
F1_{\text{BERT}} &= \frac{2 \cdot P_{\text{BERT}} \cdot R_{\text{BERT}}}{P_{\text{BERT}} + R_{\text{BERT}}}.
\end{align*}

Where $R_{\text{BERT}}$ is the recall, and $P_{\text{BERT}}$ is the precision.

\begin{figure}[ht]
\centering
  \includegraphics[width=0.99\textwidth]{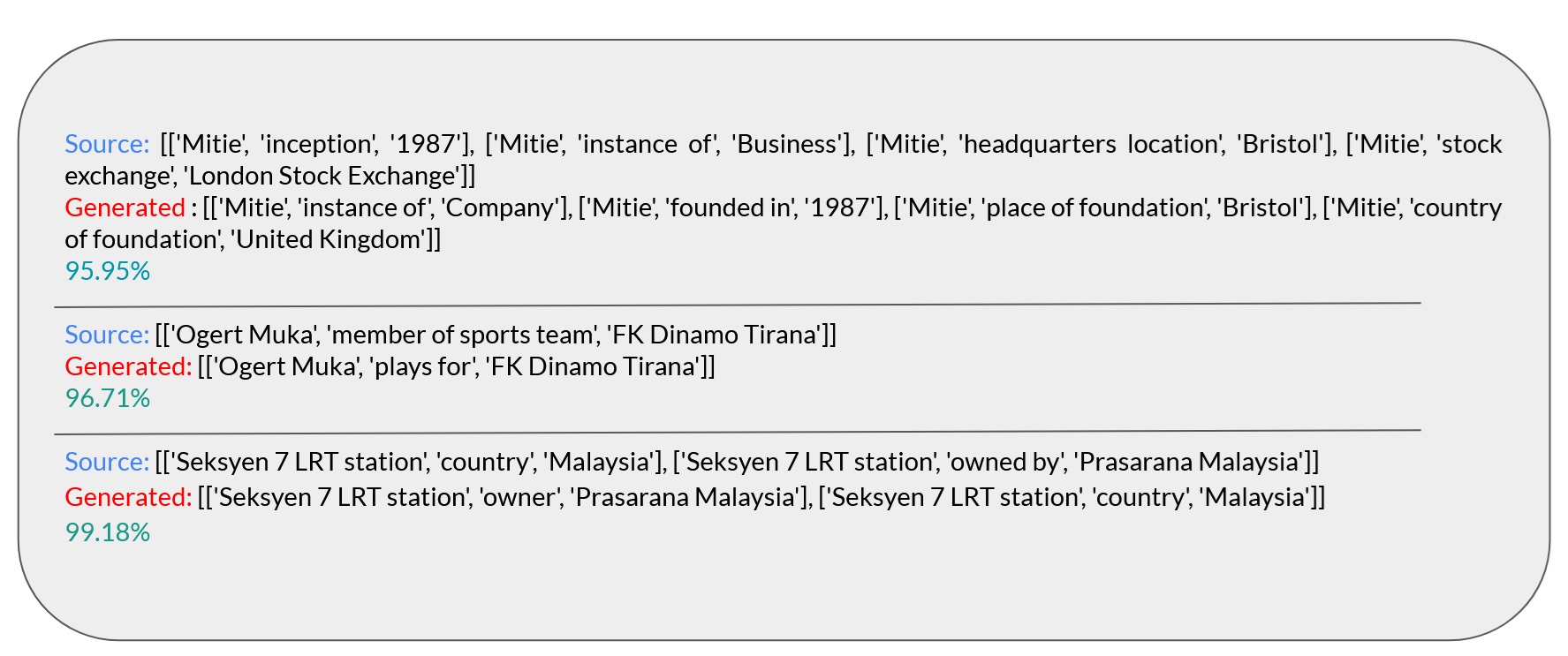}
\caption{Examples of the calculated GM-GBS}
\label{fig:4}       
\end{figure}

In this work, we use G-BS to compare generated graphs with ground-truth graphs, defining graph matching with a similarity threshold of 95\% to introduce GM-GBS. This approach acknowledges that entities or relations in the generated graph may be synonymous with those in the ground truth graph. Results shown in Fig.\ref{fig:4} illustrate that even with 95\% BERTScore similarity, the generated graph is nearly identical to the ground truth.

To calculate GM-GBS, we follow these steps: Given an array of F1 scores of G-BS \( f_1, f_2, \ldots, f_n \) in \texttt{f1s\_BS}, the fraction of F1 scores greater than 0.95 is calculated as follows:\\

1. Let $ToGrs$ be the total number of generated graphs.

2. Let \( f_m \) be the count of F1 scores that are greater than 0.95:
\begin{align*}
\bm{f_m = \sum_{i=1}^{N} 1(f_i > 0.95)}
\end{align*}
where \( 1(\cdot) \) is the indicator function, which is 1 if the condition inside is true and 0 otherwise.

3. The fraction of F1 scores greater than 0.95 is given by: \( \text{GM-GBS} = \frac{f_m}{N} \)
\\
\\
For hallucinations and omissions, we use Optimal Edit Paths (OEP) to determine exact counts:

Hallucination: An edit operation is a hallucination if it adds an entity or relation not present in the gold graph. We previously used an overall hallucination metric \( \bm{Hall. = \frac{hall}{ToGrs}} \), where $hall$ is the number of graphs with hallucinations.\\

Omission: An edit operation is an omission if it deletes an entity or relation present in the gold graph. In the previous work the omission was computed by \(\bm{Omis. = \frac{omiss}{ToGrs}}\), where $omiss$ is the number of graphs with omissions. \\

In this work, we calculate exact percentages of hallucination and omission through qualitative evaluation. \\
Given a list of tuples \( \text{lst} = [(g_1, p_1), (g_2, p_2), \ldots, (g_n, p_n)] \), where \( g_i \) represents a gold edge and \( p_i \) represents a predicted edge:

1. Let \( h \) be the number of hallucinations, where a hallucination is defined as \( g_i = \text{None} \): \\
\begin{align*}
    h = \sum_{i=1}^{n} 1(g_i = \text{None})   
\end{align*}

2. The exact hallucination rate is then calculated as: \( \text{\textbf{Hall\_Rate}} = \bm{\frac{h}{n}} \)\\

Where $n$ is the total number of edges, and \( 1(\cdot) \) is the indicator function, which is 1 if the condition inside is true and 0 otherwise (Same for Omis\_rate).\\

To calculate the exact omission rate: \\

1. Let \( o \) be the number of omissions, where an omission is defined as \( p_i = \text{None} \):
\begin{align*}
o = \sum_{i=1}^{n} 1(p_i = \text{None})\\
\end{align*}

2. The exact omission rate is then calculated as: \( \text{\textbf{Omis\_Rate}} = \bm{\frac{o}{n}} \)\\

\section{Experiments}
\label{sec:4}

This section outlines the LLMs used in our experiments for ZSP and FSP and presents the experimental results.

We utilized the Mistral model from HuggingFace platform\footnote{Hugging Face: \url{https://huggingface.co/}}, specifically focusing on the finetuned Mistral model which showed the best results in our previous work. We also compared the finetuned model with the original Mistral model.

\begin{itemize}
    \item Original Mistral-7B-v0.1: A pretrained generative text model with 7 billion parameters introduced by \cite{jiang2023mistral}, which outperforms Llama 2 13B in various benchmarks.
    \item Fine-tuned Mistral-7B-v0.1: Based on the original Mistral and fine-tuned on the WebNLG+2020 training dataset, this model outperformed other finetuned models like Llama2 (7b and 13b) and Starling in our previous work \cite{nous}.
\end{itemize}

Our evaluation also considers hallucination and omission through a linguistic lens, unlike most studies which focus on precision, recall, F1 score, triple matching, or graph matching, except for \cite{mihindukulasooriya2023text2kgbench} which includes hallucination evaluation.

Table \ref{tab:tab_res} shows that the fine-tuned Mistral performs better in both ZSP and FSP compared to the original Mistral for the T2KG construction task. The performance improves with more examples (more shots), with both finetuned and original Mistral models. Seeing the fine-tuned Mistral, it has the best performance when given 7 shots, surpassing the original Mistral by a significant margin.

As mentioned in our previous work, to corroborate these findings, in this version of our study, we assess our fine-tuned models using KELM-sub dataset for few-shot. We see that even when we gave Mistral examples from KELM-sub, it works better than zero-shot for the test dataset of WebNLG.

As depicted in Fig. ~\ref{fig:1}, Hall. represents Hallucinations, while Omis. denotes Omissions.

\begin{table*}
\caption{Comparison of performance metrics and models on WebNLG test dataset. Lower values indicate better performance for GED, Hall., and Omis.}
  \label{tab:tab_res}
  \begin{tabular}{ c | | c | c | c | c | | c | c | c } 
  \toprule
  Model | Metric & G-F1 & T-F1 & G-BS & GED & Hall. & Omis. & GM-GBS \\ 
  \midrule \midrule  
    Mistral-0 & 2.30 & 3.27 & 77.87 & 15.84 & 20.35 & 31.31 & 33.27 \\
    Mistral-7 & 18.72 & 28.44 & 87.54 & 10.13 & 17.88 & 21.14 & 51.88 \\
    \midrule
    Mistral-FT-0 & 31.93 & 44.08 & 86.89 & 8.25 & \textbf{13.55} & 18.27 & 54.97 \\
    Mistral-FT-7 & \textbf{34.68} & \textbf{49.11} & \textbf{91.99} & \textbf{6.69} & 14.90 & 14.39 & 57.72 \\ 
    \midrule \midrule     
    Mistral-6 (KELM-sub) & 7.59 & 12.45 & 81.23 & 16.29 & 61.16 & \textbf{7.64} & 26.86 \\
    Mistral-FT-6 (KELM-sub) & 31.37 & 47.49 & 91.27 & 7.51 & 27.37 & 8.26 & \textbf{58.40} \\

  \bottomrule
\end{tabular}
\end{table*}

The G-BS consistently remains high, indicating that LLMs frequently generate text with words (entities or relations) very similar to those in the ground truth graphs, which was one reason to use it for the GM-GBS metric. The finetuned Mistral with 7 shots achieves the highest G-F1, accurately generating approximately 35\% of graphs identical to the ground truth. This model performs exceptionally well across various metrics, particularly in T-F1. Additionally, the finetuned Mistral with 6 examples from KELM-sub outperforms the finetuned Mistral with 7 examples from WebNLG+2020 using the GM-GBS metric.

In Table \ref{tab:tab_res_kelm}, we present the evaluation results of the original Mistral with 7-shot learning (using examples from WebNLG+2020) and the fine-tuned Mistral with zero-shot (Mistral-FT-0) and 7-shot (Mistral-FT-7) learning (also using examples from WebNLG+2020) on the KELM-sub test dataset, prepared by \cite{han2023pive} and based on \cite{agarwal2020knowledge}. It is important to note that the experiments utilized the same prompts as previously described. The 7-shots experiments used examples from the WebNLG+2020 training dataset. These experiments aim to assess the generalization ability of the original LLMs with 7-shot learning and the fine-tuned LLMs with zero-shot and 7-shot learning across diverse domains in the T2KG construction task.

Another experiment was conducted using 6 random examples  from the KELM-sub training dataset. We applied this prompt to both the original Mistral (Mistral-6) and our finetuned Mistral (Mistral-FT-6) models. As expected, Mistral-6 outperformed Mistral-7 because the examples were from the KELM-sub training dataset used in Mistral-6. However, it was interesting to observe that Mistral-FT-6 performed less effectively than Mistral-6 with the same examples. This suggests that finetuning on WebNLG domains reduces the generalizability of the LLMs.  

The results in Table \ref{tab:tab_res_kelm} indicate that the fine-tuned Mistral models perform less effectively than the original Mistral with 7 shots from WebNLG+2020 and with 6 shots from KELM-sub. Additionally, all fine-tuned versions of Mistral (Mistral-FT-7, Mistral-FT-0, and Mistral-FT-6) show inferior results on KELM-sub compared to WebNLG+2020. This disparity can be attributed to the presence of different relation types, with some types expressed differently in KELM-sub. To address this, we utilize G-BS to calculate the similarity between two graphs and consider them as synonyms if they are sufficiently similar (\textgreater95\% of similarity). This metric, called GM-GBS (Graph Matching using Graph BERTScore), is the last metric presented in Table \ref{tab:tab_res_kelm}. GM-GBS indicates a higher value of graph matching. To assess the reliability of this metric, we conducted a qualitative evaluation as illustrated in Fig. \ref{fig:4}.

Overall, unlike our previous work where we used examples from WebNLG with the original and fine-tuned models for few-shot learning, using examples from KELM-sub here shows that the results are relatively similar. This indicates that fine-tuning negatively affects the generalization capability of the models.

\begin{table*}
  \caption{Results on KELM-sub. Lower values indicate better performance for GED, Hall., and Omis.}
  \label{tab:tab_res_kelm}
  \begin{tabular}{ c | | c | c | c |  c | | c | c | c } 
    \toprule
     Model | Metric & G-F1 & T-F1 & G-BS & GED & Hall. & Omis. & GM-GBS \\ 
     \midrule\midrule   
     Mistral-7 & 5.50 & 11.35 & 81.77 & 13.74 & 6.72 & 61.09 & 28.66 \\
     \midrule 
    Mistral-FT-0 & 2.17 & 8.55 & 78.29 & 14.35 & 7.22 & 56.28 & 12.88 \\
    Mistral-FT-7 & 2.89 & 9.92 & 78.42 & 13.63 & \textbf{6.22} & 61.00 & 13.66 \\ 
    \midrule\midrule  
    Mistral-6 (KELM-sub) & \textbf{12.00} & \textbf{31.08} & \textbf{85.49} & \textbf{10.82} & 25.50 & \textbf{32.44} & \textbf{38.88} \\ 
    Mistral-FT-6 (KELM-sub) & 4.00 & 17.66 & 84.30 & 12.50 & 11.06 & 48.17 & 36.22 \\ 
    
  \bottomrule
\end{tabular}
\end{table*}

\paragraph{Qualitative results : }
As illustrated in Fig. \ref{fig:3}, our metric precisely calculates the percentage of hallucinations and omissions in the generated graphs at the triple level. For example, if a generated graph contains 2 triples and 1 of them are not present in the ground truth graph, the hallucination rate is approximately 50\%. Similarly, for omissions, if the generated graph is missing some triples present in the ground truth graph, the omission rate is calculated accordingly.

As previously mentioned, we use G-BS to calculate the similarity between generated and ground truth graphs. If the similarity value exceeds 95\%, we consider it an exact match, based on the notion that entities or relations in the generated graph are very close to those in the ground truth graph, or what we refer to as synonyms. In Fig. \ref{fig:4}, we present examples with varying levels of similarity, including one with approximately 95\% similarity, to demonstrate that even with 95\% similarity, the two graphs convey the same or very similar meanings.

\section{Conclusion and perspectives}
\label{sec:5}
In this study, we evaluated the performance of both the original and fine-tuned Mistral models for Text-to-Knowledge Graph (T2KG) construction tasks using Zero-Shot Prompting (ZSP) and Few-Shot Prompting (FSP). Our analysis incorporated a comprehensive set of metrics, including G-F1, T-F1, G-BS, GED, along with measures for hallucinations and omissions.

Our results demonstrate that the fine-tuned Mistral model generally outperforms the original Mistral, particularly in Few-Shot scenarios. The fine-tuned Mistral with seven shots achieved superior performance across most metrics, notably improving G-F1 and T-F1 scores, which indicates a higher fidelity in generating ground truth graphs, and reflects its improved ability to produce coherent and contextually relevant outputs.

Despite these improvements, we observed that fine-tuning on domain-specific data, such as WebNLG, can negatively impact the model’s generalization capabilities. This was evident from the comparative performance of the fine-tuned models on the KELM-sub dataset, where the original Mistral model with 7 shots from WebNLG+2020 outperformed the fine-tuned variants. This finding highlights the importance of balancing domain-specific fine-tuning with maintaining broad generalization.

The inclusion of the GM-GBS metric provided valuable insights into the semantic similarity between generated and ground truth graphs. Our qualitative analysis of hallucinations and omissions further enhanced our understanding of model performance at the triple level.

Looking ahead, there are several promising avenues for further research. Refining evaluation metrics to account for synonyms of entities or relations in generated graphs could improve assessment accuracy. Additionally, leveraging LLMs for data augmentation in T2KG construction shows potential, as our experiments suggest that LLMs can maintain consistency in generating results and propose relevant triples.

Expanding evaluations to a broader range of domains and datasets can provide deeper insights into how various types of data influence model behavior and performance. Combining automated metrics with human evaluation could also offer a richer understanding of model quality, with domain experts providing valuable assessments of the relevance and accuracy of generated graphs. Exploring these directions will contribute to advancing the field of T2KG construction and enhancing the capabilities of language models in producing accurate and contextually appropriate knowledge graphs.

\section{Acknowledgments}
The authors thank the French company DAVI (Davi The Humanizers, Puteaux,
France) for their support, and the French government for the plan France Relance funding.

%
%

\end{document}